\pdfoutput=1
\documentclass[11pt]{article}

\usepackage{acl}

\usepackage{times}
\usepackage{latexsym}
\usepackage{color,soul}
\usepackage{graphicx}
\usepackage{enumitem}
\usepackage{booktabs}
\usepackage{multirow, makecell}
\usepackage{tikz}
\usepackage{graphicx}
\usepackage{makecell}
\usepackage{comment}
\usepackage{subfigure}

\usepackage[T1]{fontenc}
\usepackage[utf8]{inputenc}

\usepackage{microtype}

\usepackage{pifont}
\newcommand{\cmark}{\ding{51}}
\newcommand{\xmark}{\ding{55}}

\newcommand{\PreserveBackslash}[1]{\let\temp=\\#1\let\\=\temp}
\newcolumntype{C}[1]{>{\PreserveBackslash\centering}p{#1}}
\newcolumntype{R}[1]{>{\PreserveBackslash\raggedleft}p{#1}}
\newcolumntype{L}[1]{>{\PreserveBackslash\raggedright}p{#1}}

\newlength{\mysize}
\newcommand{\mycfs}[1]{\setlength{\mysize}{#1pt}%
  \fontsize{\mysize}{1.2\mysize}\selectfont}

\newif\ifcomments
\ifcomments
    \providecommand{\eric}[1]{{\protect\color{magenta}{[EW: #1]}}}
    \providecommand{\robin}[1]{{\protect\color{magenta}{[RJ: #1]}}}
    \providecommand{\adina}[1]{{\protect\color{magenta}{[AW: #1]}}}
    \providecommand{\douwe}[1]{{\protect\color{magenta}{[DK: #1]}}}
\else
    \providecommand{\eric}[1]{}
    \providecommand{\robin}[1]{}
    \providecommand{\adina}[1]{}
    \providecommand{\douwe}[1]{}
\fi

\title{
Analyzing Dynamic Adversarial Training Data in the Limit
}

\newcommand*\samethanks[1][\value{footnote}]{\footnotemark[#1]}
\author{
    \bf Eric Wallace$^{1}$\thanks{~~Work done while an intern at Facebook AI Research.} \hspace{0.3cm}
    \bf Adina Williams$^{2}$\thanks{~~Equal contribution} \hspace{0.4cm}
    \bf Robin Jia$^{2,3}$\samethanks \hspace{0.4cm}
    \bf Douwe Kiela$^{2}$\samethanks \hspace{0.4cm} \\
    $^1$University of California, Berkeley \hspace{0.3cm}
    $^2$Facebook AI Research \hspace{0.3em}
    $^3$University of Southern California \hspace{0.3em}\\
    \href{mailto:ericwallace@berkeley.edu}{\tt ericwallace@berkeley.edu} \hspace{0.4cm}
    \href{mailto:dkiela@fb.com}{\tt  dkiela@fb.com} \\
}

\begin{document}
\maketitle
\begin{abstract}
To create models that are robust across a wide range of test inputs, training datasets should include diverse examples that span numerous phenomena.
Dynamic adversarial data collection (DADC), where annotators craft examples that challenge continually improving models, holds promise as an approach for generating such diverse training sets.
Prior work has shown that running DADC over 1--3 rounds can help models fix some error types, but it does not necessarily lead to better generalization beyond adversarial test data.
We argue that running DADC over \textit{many} rounds maximizes its training-time benefits, as the different rounds can together cover many of the task-relevant phenomena. 
We present the first study of longer-term DADC, where we collect 20 rounds of NLI examples for a small set of premise paragraphs, with both adversarial and non-adversarial approaches.
Models trained on DADC examples make 26\% fewer errors on our expert-curated test set compared to models trained on non-adversarial data.
Our analysis shows that DADC yields examples that are more difficult, more lexically and syntactically diverse, and contain fewer annotation artifacts compared to non-adversarial examples.
\end{abstract}

\section{Introduction}

\begin{figure}[t]
\centering
\includegraphics[trim={0.0cm 0cm 0.0cm 0.0cm},clip, width=1.0\columnwidth]{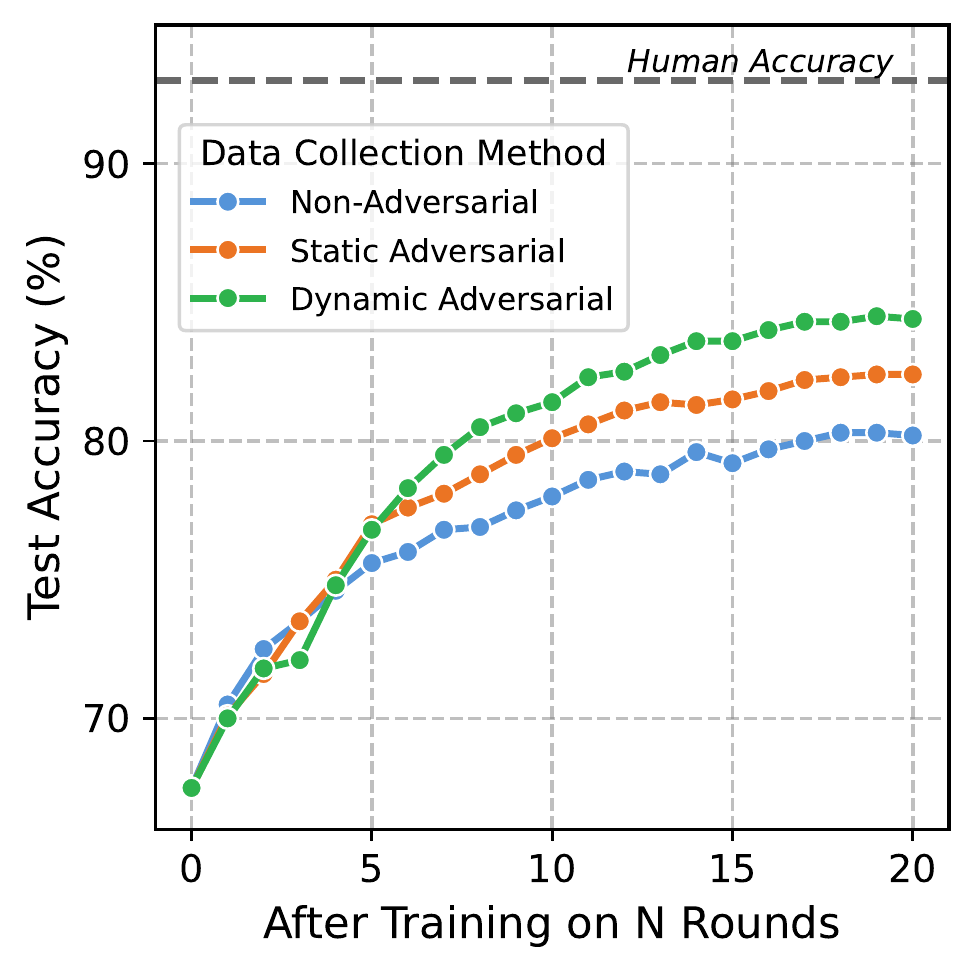}
\vspace{-0.8cm}
\caption{\textit{Model accuracy on our expert-curated test set} when training on data collected from three different methods. Standard non-adversarial data collection is more effective than adversarial data collection in the short-term. However, in the long term, adversarial data collection statistically significantly outperforms standard data, especially when the data is collected using a dynamic model that is updated after each round.}
\label{fig:test}
\end{figure}

Traditional crowdsourcing methods often yield datasets that
lack diversity, contain spurious correlations, and are easy for existing models~\cite{gururangan-etal-2018-annotation,poliak-etal-2018-hypothesis,geva-etal-2019-modeling,ko2020look,potts-etal-2021-dynasent}. Training on such examples can lead to models that reach deceptively high accuracy on in-distribution test data, yet fail on challenge sets~\cite{naik-etal-2018-stress,glockner-etal-2018-breaking,gardner-etal-2020-evaluating}, input perturbations~\cite{wallace-etal-2019-trick,kaushik-etal-2020-learning}, and distribution shifts~\cite{talmor-berant-2019-multiqa,hendrycks-etal-2020-pretrained}.

Dynamic adversarial data collection (DADC) holds promise as an approach to mitigate these training set problems. In DADC, humans are tasked with creating examples that fool state-of-the-art models but are answerable by humans. Crucially, DADC is dynamic in that data collection is repeated over many rounds with a stream of ever-improving models-in-the-loop. As models improve, annotators are incentivized to craft new types of examples that challenge the latest models. In the limit, this process would ideally cover most task-relevant phenomena, leading to more robust models.

Whether DADC actually leads to diverse, high-coverage training data, however, has remained unclear. 
It could cause annotators to write unnatural examples or to focus on a narrow subset of unusual examples that models find difficult to learn, thus decreasing data diversity~\cite{bowman-dahl-2021-will}.
Some prior work has shown that a few rounds of DADC can indeed improve robustness to adversarial inputs \citep{dinan-etal-2019-build,nie-etal-2020-adversarial},
however, there are mixed results on improving accuracy on other distributions~\cite{kaushik-etal-2021-efficacy}.
To date, no study has analyzed how DADC evolves over \textit{many} rounds. Thus, the long-term benefits or drawbacks of adopting it as a core dataset creation paradigm remain poorly understood.

In this work, we conduct the first study of DADC's effects in the long term, where we conduct many rounds and rapidly update models. We focus on the task of natural language inference (NLI), which serves as a crucial benchmark for research on language understanding~\cite{bowman-etal-2015-large,williams-etal-2018-broad}. To make our study feasible, we conduct intensive data collection on a small set of context passages that span different genres and exhibit numerous natural language phenomena.
By using a small set of contexts, we create a scenario in which models can improve quickly from round to round, thus approximating the dynamics of running DADC at a larger scale.
We compare three approaches for collecting training data---no model, static model-in-the-loop, and dynamic model-in-the-loop---in a controlled setting for 20 rounds.

To evaluate the different methods, we collect expert-curated non-adversarial test examples for each context that span numerous NLI phenomena which humans can handle correctly.
On this test set, DADC outperforms the alternative approaches after many rounds of data collection (e.g., Figure~\ref{fig:test}). Standard non-adversarial data collection causes model accuracy to climb quickly for a short period of time, but accuracy quickly plateaus after more examples are collected. On the other hand, DADC examples yield larger improvements for later rounds. To understand these results, we show that DADC examples are overall more diverse in lexical and syntactic patterns, contain fewer artifacts, and become more difficult over each round. 
Overall, our results show that building large adversarial training sets may be more useful than standard model-agnostic collection in the long term.
\section{Background}

\noindent \textbf{Collecting Data with Crowdsourcing.} Most large-scale supervised datasets are collected using crowd workers~\cite{bowman-etal-2015-large,rajpurkar-etal-2016-squad, kocisky-etal-2018-narrativeqa}. Compared to experts, crowd workers often produce lower quality data as they are not necessarily well-trained for one's task and can be apathetic to the goals of the research~\cite{snow-etal-2008-cheap,ujwal2017turk}.
These data quality issues are exacerbated for language tasks because crowd workers also need to \textit{write} inputs, e.g., writing hypothesis sentences for natural language inference tasks. These manually-written inputs often follow a very narrow distribution: they lack diversity over lexical items, syntactic patterns, domains, example difficulties, reasoning types, and more~\cite{yang-etal-2018-mastering,gururangan-etal-2018-annotation,geva-etal-2019-modeling,min-etal-2019-compositional,kiela-etal-2021-dynabench}. 

\paragraph{Dynamic Adversarial Data Collection.} In DADC, workers are tasked with writing examples that are answerable by humans but fool existing models~\cite{wallace-etal-2019-trick,nie-etal-2020-adversarial,kiela-etal-2021-dynabench}. Concretely, workers are presented with a user interface where they can observe model predictions and interactively build data that exposes model failures. Multiple \emph{rounds} may also be conducted, where the model is updated on the adversarial data collected thus far and redeployed; the goal of this is to encourage workers to write increasingly more difficult examples.
Adversarial data collection has been widely adopted in recent work, especially for building \textit{evaluation} datasets~\cite{dua-etal-2019-drop,nie-etal-2020-adversarial,dinan-etal-2019-build,bartolo-etal-2020-beat,potts-etal-2021-dynasent,liu2020nyt,kaushik-etal-2021-efficacy, xu-etal-2020-recipes, xu-etal-2021-bot}. Our focus is instead on \emph{training}, where past work has shown that after a few rounds of adversarial data, a model noticeably improves on its errors, yet many problems still remain~\cite{nie-etal-2020-adversarial,bartolo-etal-2020-beat,kaushik-etal-2021-efficacy,zellers-etal-2019-hellaswag}. Moreover, it remains unclear whether collecting adversarial or non-adversarial data leads to generally more robust models in the long term~\cite{kaushik-etal-2021-efficacy}.

\begin{table*}[!t]
\centering
\footnotesize
\begin{tabular}{C{1cm}lcp{8.65cm}ll}
\toprule
\textbf{Premise} & \textbf{Model} & \textbf{Rd} & \textbf{Hypotheses} & \textbf{Label} & \textbf{Error} \\
\midrule
\multirow{3.5}{*}{Sound} & No & 20 & Old telephones have sheepskin over a cup or cylinder. & Entail & -\\[0.4ex]
                            & Static & 20 & Parts of animal anatomy can function as the origins of sound. & Entail & \xmark \\[0.4ex]
                            & Dynamic & 20 & The transmission due to the vibration can be attenuated with distances. & Entail & \cmark \\
\midrule
\multirow{3.5}{*}{Yellow} & No & 20 & Ruiz's experiment was on three men. & Contradict & -\\[0.4ex]
                            & Static & 20 & It turned out that basset hounds were immune to yellow fever. & Contradict & \cmark \\[0.4ex]
                            & Dynamic & 20 & The American Public Health Association meeting, held in October 1900, was about developing vaccines against yellow fever. & Contradict & \xmark \\
\midrule
\multirow{3.5}{*}{Faraday} & No & 20 & michael faraday's mother was named margaret & Entail & -\\[0.4ex]
                            & Static & 20 & The home of the Faradays, in London, was very crowded. & Entail & \xmark \\[0.4ex]
                            & Dynamic & 20 & Michael had at least nine uncles and/or aunts. & Entail & \cmark \\
\bottomrule
\end{tabular}
\vspace{-0.2cm}
\caption{Examples from the training sets that are generated by crowd workers, with \emph{No}, \emph{Static}, or \emph{Dynamic} models in the loop. The error column shows whether the worker successfully fooled the model in the loop when submitting the example in the user interface. See Table~\ref{tab:contexts} for the full premise paragraphs.}
\label{tab:qualitative_train}
\end{table*}

\section{Dynamic Data Collection in the Limit}
The paradigm of DADC raises a natural but unanswered question: what would happen if we kept going? 
If we ran DADC for many years, how robust would the resulting models be? 
Would models improve more quickly than if we had collected training data without a model-in-the-loop?

Answering these forward-looking questions is key to understanding whether researchers and practitioners should continue to collect data in an adversarial fashion.
Of course, we cannot practically run many years of data collection at once due to cost and time constraints. 
Our key idea is to instead answer these questions for a more manageable test bed that still retains many of the key challenges associated with language understanding tasks.
In particular, we scale down the natural language inference (NLI) task to a small number of paragraph-length premises.
In this setting, many rounds of smaller-scale data collection can tell us whether DADC or non-adversarial data collection leads to more robust model accuracy on test hypotheses for these same contexts.
If DADC is indeed superior, this suggests that DADC can collect data that more effectively covers the challenging phenomena required for NLI, and therefore scaling it up to (many) more contexts could yield models that are similarly robust for more general NLI.

\subsection{Task and Context Paragraphs}
We choose to focus on NLI, a canonical and well-studied natural language understanding task \citep{dagan2005pascal,bos-markert-2005-recognising,giampiccolo-etal-2007-third,maccartney-manning-2009-extended}. 
NLI training datasets are notorious for being rife with artifacts and biases~\cite{poliak-etal-2018-hypothesis,gururangan-etal-2018-annotation, tsuchiya-2018-performance,mccoy-etal-2019-right}, which makes NLI a suitable test bed for studying questions surrounding training dataset quality.
Using NLI also enables us to write a rich and diverse test set with a small number of contexts because each premise admits many possible hypotheses.
We focus on \textit{binary} NLI---definitely entailing or not entailing---to minimize labeling disagreements stemming from the distinction between neutral and contradiction in three-way NLI~\cite{pavlick-kwiatkowski-2019-inherent, nie-etal-2020-learn}. 

We use ten diverse paragraphs from Project Gutenberg\footnote{\url{https://www.gutenberg.org/}} as the premises---each one is chosen to elicit many possible hypotheses.
We choose these paragraphs to span a range of genres (scientific, biographical, historical, narrative) and present a different set of challenges.
For instance, some passages describe physical objects in detail, requiring commonsense understanding of the physical world (e.g., \textit{``\dots Phonny had not measured his wires in respect to length, but had cut them off of various lengths, taking care however not to have any of them too short.  The result was that the ends of the wires projected to various distances above the board\dots''}).
Other passages describe reasoning about uncertainty (e.g., \textit{``\dots this negative result might be because these animals are not susceptible to the disease\dots''}) or hypothetical events (e.g., \textit{``\dots If there should be even partial cooperation between the Austrian leaders, he must retreat \dots''}).
See Appendix~\ref{appendix:contexts} for the full premise paragraphs. 
We minimally edit each paragraph so that they can be read standalone, e.g., we resolve coreferences. 

\subsection{Data Collection Procedure}\label{subsec:collection}
We collect data over many rounds, where each round comprises three steps. First, crowdworkers write hypothesis sentences that are either entailed or not entailed by one of our premises while interacting with the current model-in-the-loop. 
Second, other crowdworkers relabel these examples and help filter out spam and other malformed examples.
Finally, we update the model-in-the-loop by fine-tuning on all collected data, including data from the newest round.
We use Amazon Mechanical Turk~(AMT) for data collection. 

\paragraph{Hypothesis Generation.} To generate hypotheses, we run AMT tasks where a worker is randomly provided one of the premises and is asked to write ten different hypotheses. After writing each hypothesis, they are shown the predictions of a live model in the loop. 
To encourage workers to write model-fooling examples, they are given a bonus every time one of their examples fools the model and passes the later label verification step.
We ask workers to write ten hypotheses for a single premise, as this allows them to better understand the model's behavior and empirically leads to more-difficult examples (Section~\ref{sec:results}). 
The worker can generate hypotheses for either of the binary labels, but we encourage them to generate balanced examples in the onboarding instructions. The user interface is shown in Appendix~\ref{appendix:turk}.

\paragraph{Label Verification.} To ensure the generated hypotheses are labeled correctly, we run a separate AMT task where workers are asked to label each example without being shown the original label. Each example is labeled by at least three workers. If all three agree, that example is saved. If there is a disagreement, we ask two additional workers and keep the example if four out of five agree on the label. We also provide an option to flag a hypothesis as ``bad'', e.g., it is very ungrammatical or clearly spam. If more than one worker flags an example as bad, we remove it.
We do not allow workers to participate in both the labeling and validation AMT tasks, as we do not want workers to be influenced by one another's hypotheses. 

\paragraph{Updating the Model.}
For the initial round of data collection, we use as our starting point a RoBERTa-large model~\cite{liu2019roberta} that has been finetuned on SNLI~\cite{bowman-etal-2015-large}, MNLI~\cite{williams-etal-2018-broad}, and FEVER-NLI~\cite{nie2019combining}.%\footnote{FEVER-NLI builds upon the dataset from the FEVER shared task~\cite{thorne-etal-2018-fever}. The SNLI and FEVER datasets are \href{https://creativecommons.org/licenses/by-sa/4.0/}{licensed CC-BY-SA}. MNLI is \href{https://github.com/nyu-mll/multiNLI/blob/master/LICENSE.txt}{licensed by MIT.} and its copyright is held by New York University (2018).} 
We use this training data as it provides us with an accurate initial model, and note that we collapse the neutral and contradiction labels during training as we focus on binary NLI.
To update the model after each round, we continue finetuning it on all of the data collected thus far and then deploy it for the next round. Our finetuning hyperparameters follow the recommendations of \citet{mosbach-etal-2021-stability}: we use a learning rate of $2 \times 10^{-5}$, a learning rate warmup over the first 10\% of steps, bias-corrected Adam, and 15 epochs of training. We early stop using held-out validation data (see Section~\ref{subsec:details}). We refer to this setting, where crowdworkers interact with a model-in-the-loop that is updated after each round, as the \textbf{Dynamic Model} setting.

\paragraph{Baselines.} In addition to the above, we also collect data with two baseline approaches:

\begin{itemize}[itemsep=0mm,topsep=3pt,leftmargin=10pt]
\item \textbf{No Model.} This is the typical procedure for collecting training data where workers do not interact with a model.
\item \textbf{Static Model.} We provide a model in the loop to the workers but the model is kept fixed across all the rounds.
We use the same model that the Dynamic Model setting uses in its first round.
\end{itemize}
No data is mixed between methods and workers can not participate in multiple methods.

\subsection{Dataset Details}\label{subsec:details}
Our codebase is built on top of the Dynabench platform~\cite{kiela-etal-2021-dynabench}, we deploy tasks using the Mephisto library,\footnote{\url{https://github.com/facebookresearch/mephisto}} and we serve models using Dynalab~\cite{ma2021dynaboard}.
We restrict our AMT workers to those that speak English, have completed at least 100 tasks on AMT, and have an approval rating of at least 97\%. To qualify for the task, a worker must also pass an onboarding procedure where they are tasked with correctly labeling five NLI examples in a row.

For each data collection method, we run 20 rounds of data collection. We stop at 20 rounds as model performance on our validation sets begins to saturate. We collect 550 examples per round before label verification, with an equal distribution over the ten premises. All the data collection methods are run in parallel at the same time of day to control for the effects of time on data quality~\cite{karpinska2021perils}. 
At this scale, we are able to complete each round of data collection for all three methods in approximately 24 hours.
We hold out 50 examples from each round to use for early stopping and for reporting validation metrics.

\begin{figure}[t]
\centering
  \centering
  \includegraphics[width=\linewidth]{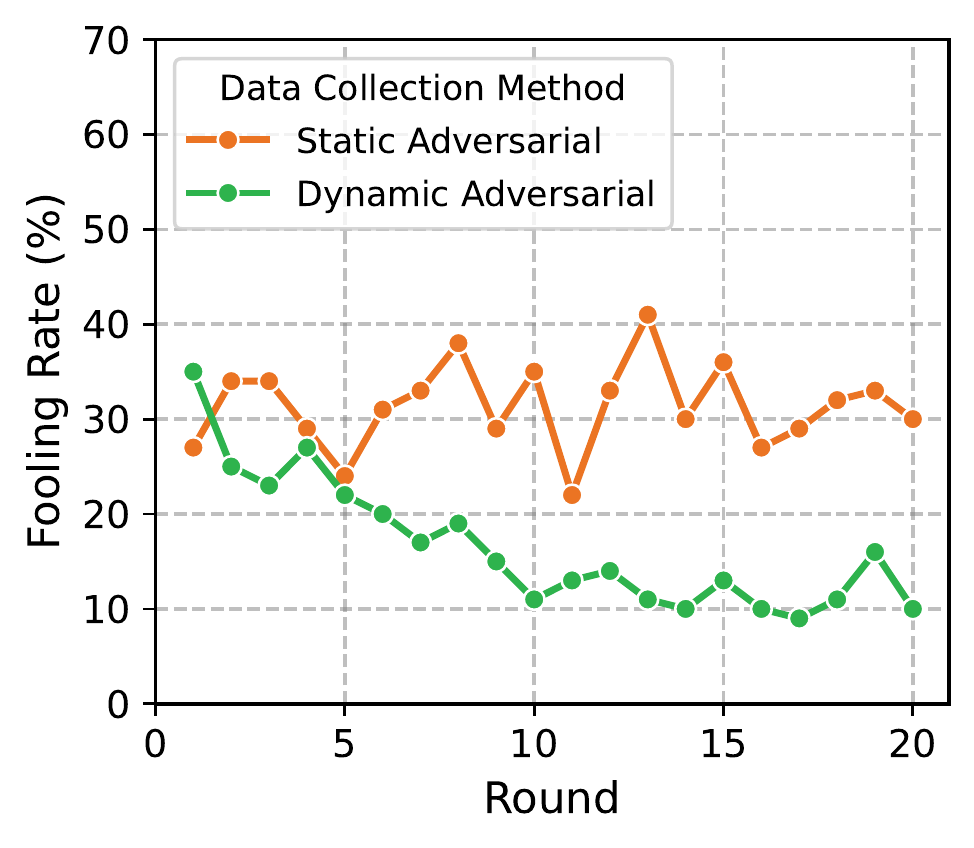}
  \vspace{-0.85cm}
  \caption{\textit{Model fooling rates.} We show how often crowd workers write examples that are successfully answered by humans but fool the model they interact with. For the static model, the fooling rate is relatively constant as the model is kept fixed (the variance across rounds is due to different crowd workers having different fooling rates). For the dynamic model, the fooling rate goes down over time as the model is updated.}
  \label{fig:fooling}
\end{figure}

Table~\ref{tab:statistics} shows overall statistics of our final datasets. These statistics are similar across the three datasets, including the label balance, the rate at which examples are discarded, and the number of AMT workers. However, the datasets differ in the rate at which workers fooled the models in the loop; Figure~\ref{fig:fooling} shows that the fooling rate is relatively constant for the static model but goes down for the dynamic model as the model is updated. Table~\ref{tab:qualitative_train} shows qualitative examples of training hypotheses from each method. We release our data and models.\footnote{\url{https://github.com/facebookresearch/dadc-limit}}

\begin{table}[t]
\begin{center}
\begin{tabular}{lccc}
 & \thead{\textbf{No} \\ \textbf{Model}} & \thead{\textbf{Static} \\ \textbf{Model}} & \thead{\textbf{Dynamic} \\ \textbf{Model}} \\[-0.5ex]
\midrule
\# Rounds & 20 & 20 & 20\\
\# Hypo. & 11,000 & 11,000 & 11,000 \\
\# Verified Hypo. & 7,684 & 7,102 & 6,911 \\
\# Workers & 115 & 104 & 121 \\
\% Contradiction & 58.5 & 56.3 & 54.6 \\
\bottomrule
\end{tabular}
\end{center}
\vspace{-0.3cm}
\caption{\textit{Statistics of our datasets.} For each method, we independently run 20 rounds of data collection with 550 hypotheses per round. We verify the labels of each hypothesis using additional crowd workers and discard any low-agreement examples; the adversarial data is discarded slightly more often. The datasets are roughly balanced between entailment and contradiction.}
\label{tab:statistics}
\end{table}

\subsection{Expert-Curated Test Set}
\label{sec:expert}
\citet{kaushik-etal-2021-efficacy} compared standard data collection to a single round of adversarial data collection, finding that adversarial training data improves accuracy only on adversarially-constructed test datasets but not on others. 
We hypothesize that running DADC for many rounds can overcome this limitation and improve generalization to independent, non-adversarial test data.
To test this, we built an expert-curated test set for our ten premise paragraphs that is intended to be challenging but not necessarily adversarial to models.
We (three of the authors) wrote 680 NLI examples, and we recruited five researchers who have published in NLI and spurious correlations to write an additional 320 examples. The test set spans different challenges, syntactic patterns, and reasoning types, loosely inspired by the categorizations from \citealt{williams2020anlizing}. The examples are not written with a model in the loop, they are balanced across the labels,%\footnote{Our test set is balanced but the three training sets are biased towards contradiction in slightly different amounts. We repeated our experiments by subsampling each training set to be balanced and found nearly identical results.}
~and they are equally distributed over the premises. Examples are shown in Table~\ref{tab:qualitative_test}.

We also collect crowd worker labels for our test set to ensure that the labels are unambiguous and to measure human accuracy. First, we collect 15 labels for each example. We remove any example from the test set where 9 or fewer workers chose the correct label; this removed 21 examples. Second, we collect an additional 5 labels to use for estimating human accuracy. The average accuracy is \textbf{93.2\%} when using each label individually.

\begin{table*}[!t]
\centering
\footnotesize
\begin{tabular}{C{1.2cm}p{11cm}L{2cm}}
\toprule
\textbf{Premise} & \textbf{Hypotheses} & \textbf{Label} \\
\midrule
\multirow{2.5}{*}{Sound} & The head of a drum and the strings of a piano are similar in that they both vibrate. & Entailment \\[0.4ex]
                            & A piano produces sound because the keys vibrate when they are struck by the pianist. & Contradiction \\
\midrule
\multirow{2.5}{*}{Yellow} & The speaker only ran one experiment of injecting yellow fever blood into animals. & Contradiction \\[0.4ex]
                            & Dr. Daniel Cruz took blood from a sick patient to run his experiment. & Entailment \\
\midrule
\multirow{2.5}{*}{Faraday} & Michael Faraday's wife was named Margaret Hastwell. & Contradiction \\[0.4ex]
                            & Yorkshire is a less populous locality to be from then Manchester Square. & Entailment \\
%\midrule
% \midrule
% \multirow{2.5}{*}{Weather} & Swallows are birds. & Entailment \\[0.4ex]
%                             & All signs of weather change are generally known. & Contradiction \\
% \midrule
% \multirow{2.5}{*}{Battle} & Alvinczy and Massena fought on the French side of the conflict. & Contradiction \\[0.4ex]
%                             & Armies have at most two flanks. & Entailment \\
% \midrule
% \multirow{2.5}{*}{Tariff} & The South imported goods from New England. & Contradiction \\[0.4ex]
%                             & Federalist merchants used to be unhappy with the government's policies. & Entailment \\
% \midrule
% \multirow{2.5}{*}{Water} & The cows are on the same side of the river as Tony. & Contradiction \\[0.4ex]
%                             & Tony could not cross the river. & Entailment \\
% \midrule
% \multirow{2.5}{*}{Wires} & Phonny had a ragged and unworkmanlike appearance. & Contradiction \\[0.4ex]
%                             & Wallace is critical of Phonny’s work. & Entailment \\
% \midrule
% \multirow{2.5}{*}{Garrity} & Garrity shot players who showed weakness. & Contradiction \\[0.4ex]
%                             & The Rockets were an organized baseball team that was not part of the minor league. & Entailment \\
\bottomrule
\end{tabular}
\vspace{-0.2cm}
\caption{Examples from our expert-curated test set. See Table~\ref{tab:contexts} for the premise paragraphs.}
\label{tab:qualitative_test}
\end{table*}
\section{Dynamic Adversarial Data Outperforms Non-Adversarial Data}\label{sec:results}

Here, we show that DADC outperforms both standard and static adversarial data collection in the long term. In particular, we train various models using the three different datasets and compare them on the validation and expert-curated test sets. 

\subsection{Training Final Models}
For each dataset, we train 20 models---one for each round---on all of the training data up to and including a given round. All models start with the same RoBERTa-large model that was used for round one of adversarial data collection. We then continue finetuning this model on the associated training data using the hyperparameters from Section~\ref{subsec:collection}. Moreover, to measure possible variance across different finetuning runs, we train each model with five different random seeds.

\subsection{Main Results} Figure~\ref{fig:test} shows our models' accuracy on the expert test set described in Section~\ref{sec:expert}. In the short term, standard non-adversarial data collection performs best---it has the highest accuracy after the first four rounds. However, in the long term, adversarial data collection, especially when done dynamically, leads to the highest accuracy by a noticeable margin. We run McNemar’s statistical test to compute whether the results are significantly different for the final round 20 models: the DADC model outperforms the static adversarial model with $p<0.05$ and the non-adversarial model with $p<0.01$; the static adversarial model outperforms the non-adversarial model with $p<0.05$.

\begin{figure}[t]
\centering
\includegraphics[trim={0.0cm 0cm 0.0cm 0.0cm},clip, width=1.0\columnwidth]{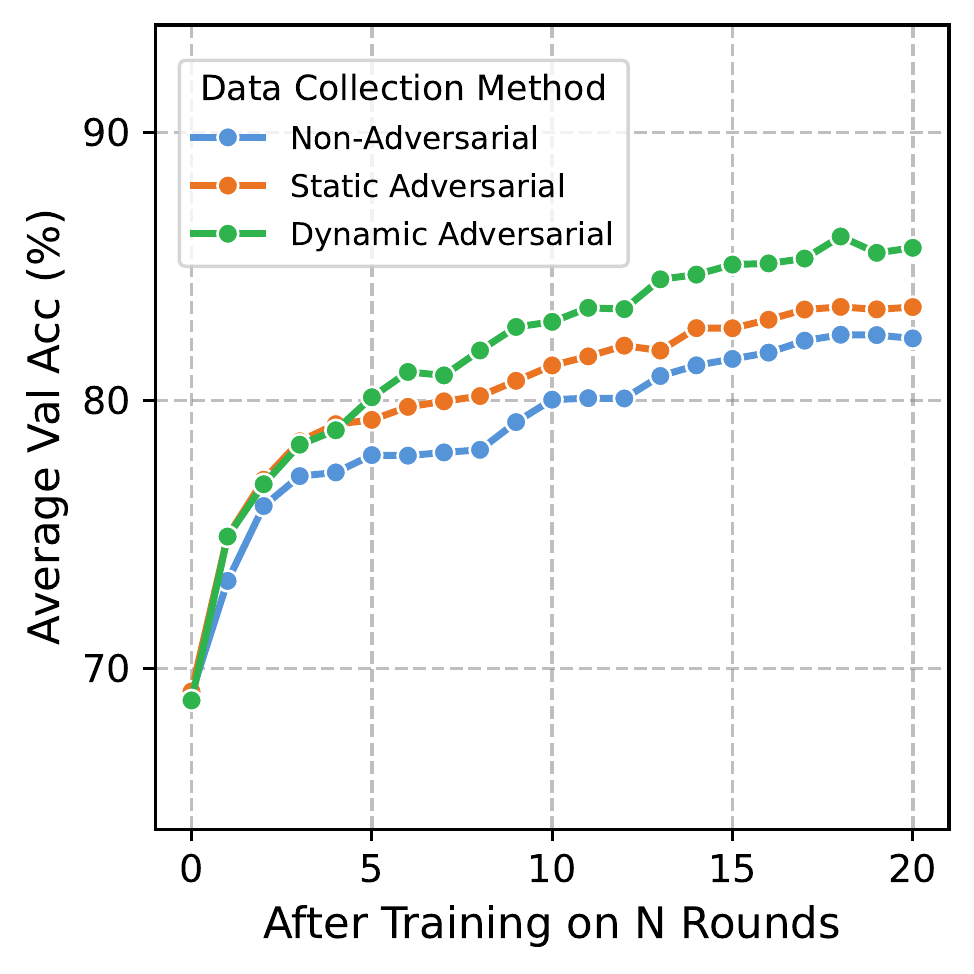}
\vspace{-0.8cm}
\caption{\emph{Combined validation accuracy}. We create a validation set by pooling together validation data from each data collection method. We find the same trend as the expert-curated test set---dynamic adversarial data performs best in the long term.}
\label{fig:validation}
\end{figure}

We also evaluate models on validation data that is split off from each round of each data collection method.
Figure~\ref{fig:validation} shows results on a validation set that is created by pooling validation data from all three collection methods; we observe the same trends as our test set, although the accuracies are slightly higher on average. 
%We also report results on each individual validation set (Figure~\ref{fig:validation_acc_break} in Appendix~\ref{appendix:accuracy}). 
%When the training data collection method matches the test data collection method, e.g., \textit{no model} training set with \textit{no model} validation set, the accuracy is the highest, as expected. However, on average across the three sets, the data collected using a dynamic model yields the highest accuracy.

Overall, these results show that when building training sets in our setting, adversarial data is not necessarily preferred when the number of examples is small. On the contrary, when the number of training examples and rounds is large, using DADC leads to more robust, broader coverage models.\smallskip

\noindent \textbf{Costs of DADC.} DADC examples are more expensive to collect as it takes crowd workers longer on average to write and verify them. However, DADC examples provide more ``bang for your buck''---it is more cost-effective to collect few DADC examples compared to many regular examples. This can be seen from Figure~\ref{fig:test} while accounting for DADC being approximately two times more expensive per example than non-adversarial data.\smallskip

\noindent \textbf{Comparison to Humans.} Even though the round 20 models have approximately 700 training examples for each premise, they are still noticeably worse than human accuracy. In particular, the best DADC model reaches 84.4\% accuracy, whereas human accuracy is 93.2\%. This shows that while DADC does lead to better models, we are still far from creating NLP systems that perform robust NLI on our premise paragraphs.\smallskip

\noindent \textbf{Generalization of DADC Data Across Models.} One possible concern with adversarially-collected data is that it could be too model-specific, similar to datasets built with active learning~\cite{lowell-etal-2019-practical}. To test whether the DADC data can generalize to other (newer) models, we train an ALBERT XXLarge-v2 model~\cite{lan-etal-2020-albert} on SNLI, MNLI, and FeverNLI. We then finetune the model on the data from all 20 rounds for each of our three datasets. The model has an accuracy of 69.1\% before updating on our collected data, and it reaches an accuracy of 83.1\%, 84.6\%, and 85.8\% on the no model, static model, and dynamic model datasets, respectively. This shows that our DADC data does generalize to better models---it leads to the highest accuracy among the three datasets---but the gap from DADC to static adversarial data is smaller than one from our RoBERTa model.\smallskip

\noindent \textbf{Generalization Beyond Our Premises.} Since the DADC data is more difficult than typical crowdsourced data, it may promote models to learn more robust NLI features. To evaluate this, we test our round 20 models on out-of-distribution datasets, including HANS~\cite{mccoy-etal-2019-right} and the MNLI mismatched test set~\cite{williams2017broad}. We convert both test sets to binary classification by collapsing the neutral and contradiction labels. We found that the round 20 models from all three settings, as well as our initial model trained on SNLI, MNLI, and FEVER-NLI, reached comparable accuracies on these test sets. This shows that while the DADC data does lead to improved in-distribution test performance, it does not necessarily lead to better performance under distribution shift.
\section{Analyzing Adversarial Data}

Why is dynamic adversarial data superior to standard data in the long term? 
In Table~\ref{tab:diversity}, we report summary statistics about our three collected datasets.
We find that dynamic adversarial data is more diverse, has higher complexity, and contains fewer artifacts than non-adversarial data.
These findings agree with our intuition surrounding adversarial datasets: small adversarial training sets that contain diverse and challenging examples may be hard for models to learn from. However, larger datasets of this type will ultimately lead to more accurate and robust models in the long term.
We describe our analyses in detail below.\footnote{Note that when computing each metric, we use a version of the No Model and Static Model datasets that are randomly downsampled to be the same size as the dynamic model data (6,911 examples). This controls for any effect that dataset size would have on our analyses.}

\begin{table}[t]
\begin{center}
\begin{tabular}{llll}
 & \thead[l]{\textbf{No} \\ \textbf{Model}} & \thead[l]{\textbf{Static} \\ \textbf{Model}} & \thead[l]{\textbf{Dynamic} \\ \textbf{Model}} \\[-0.5ex]
\midrule
\multicolumn{4}{l}{\hspace{0.1cm} \emph{Diversity}} \\
Unique Unigrams & 4.0k & 4.2k & \textbf{4.3k} \\
Unique Bigrams & 23.3k & 24.8k & \textbf{25.6k} \\
Inter-example Sim. & 41.2 & 41.9 & \textbf{39.5} \\
\midrule
\multicolumn{4}{l}{\hspace{0.1cm} \emph{Complexity}} \\
Syntax & 2.0 & 2.1 & \textbf{2.3} \\
Reading Level & 4.9 & 5.4 & \textbf{5.9} \\
Length & 10.1 & 10.9 & \textbf{12.1} \\
\midrule
\multicolumn{4}{l}{\hspace{0.1cm} \emph{Artifacts}} \\
Hypo-only Acc \% & 75.4 & \textbf{69.3} & 69.7 \\
Overlap Entail \% & 54.2 & 49.2 & \textbf{47.3} \\
\bottomrule
\end{tabular}
\end{center}
\vspace{-0.3cm}
\caption{\textit{Dataset analysis.} The hypotheses generated by DADC are more diverse based on the number of lexical items and inter-example similarity. The hypotheses are also more complex, as shown by their increased syntactic complexity, reading level, and lengths. Finally, adversarial data contains fewer instances of known artifacts. We bold the best result---lower is better for inter-example similarity and the artifact analyses.}
\label{tab:diversity}
\end{table}

\paragraph{Diversity.} DADC data is more diverse at both the lexical (unigram and bigram) and example levels (Table~\ref{tab:diversity}, top).
To measure lexical diversity, we count the number of unique unigrams and bigrams in the dataset. 
%The dynamic adversarial data is more lexically diverse---it has the most unique unigrams and bigrams.
To measure example-level diversity, we iterate through each training example and find the most similar other training sample according to BLEU score~\cite{papineni-etal-2002-bleu}. We then report the average of these BLEU scores similarities; the dynamic adversarial examples are the least similar to one another.\footnote{Note that this diversity metric is effective because we collect hundreds of examples for a single context paragraph; otherwise, we would need to measure similarity between hypotheses for different premises, a more complicated problem. We also experimented with BERTScore~\cite{zhang2019bertscore} and found similar trends as BLEU score.} The difference in inter-example similarity between the DADC data and the static adversarial data is significant with $p<0.01$ according to a $t$-test.

\begin{figure}[t]
\centering
  \centering
  \includegraphics[width=\linewidth]{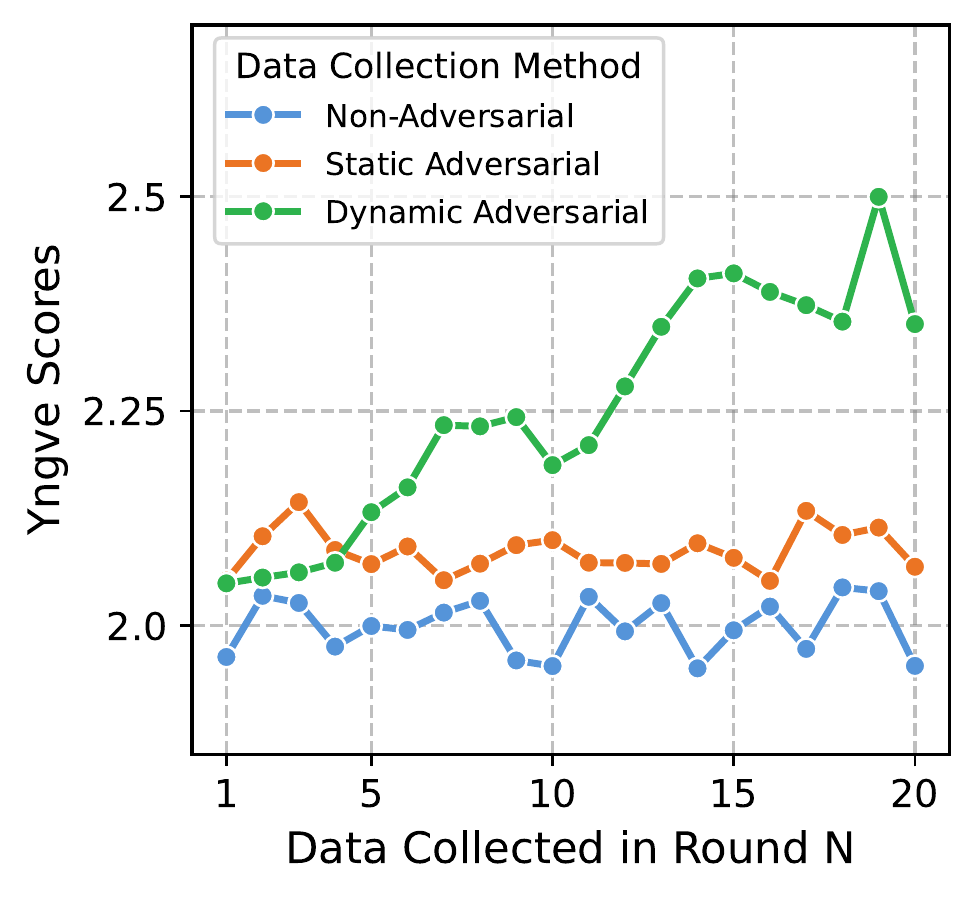}
  \vspace{-0.75cm}
  \caption{\textit{Complexity of syntax over time.} We show how the average syntactic complexity changes over each round. For the non-adversarial and static adversarial data, the syntactic complexity is relatively constant across rounds. On the other hand, the DADC examples become increasingly more complex as annotators are faced with ever-improving models in the loop.}
  \label{fig:syntax}
\end{figure}

\paragraph{Syntax and Sentence Complexity.} The dynamic adversarial data is more complex (Table~\ref{tab:diversity}, middle). For each hypothesis, we measure its length in words, its Flesch-Kincaid readability~\cite{flesch1948}, and its syntactic complexity using Yngve scores~\cite{yngve1960,roark-etal-2007-syntactic}. Yngve scores roughly measure the deviation of a parse tree from a purely right-branching tree---it is the average number of left branches on the path from the root node to each word.
To compute Yngve scores, we parse sentences using the Benepar parser~\cite{kitaev-klein-2018-constituency} based on T5 small~\cite{raffel-etal-2020-exploring}.
In all three metrics, the dynamic adversarial data scores highest, and it is statistically significantly higher than the static model data based on a $t$-test with $p<0.05$. We also show how the syntactic complexity evolves over the rounds in Figure~\ref{fig:syntax}. For the non-adversarial and static adversarial data, the syntactic complexity is relatively constant while the DADC examples become increasingly more complex.

\paragraph{Fewer Artifacts.} NLI training datasets are known to  suffer from spurious correlations. The DADC examples contain fewer instances of two known artifacts: hypothesis-only information~\cite{poliak-etal-2018-hypothesis,gururangan-etal-2018-annotation, tsuchiya-2018-performance} and high-overlap entailment examples~\cite{mccoy-etal-2019-right}. To measure such artifacts, we first train a hypothesis-only model on the training set for each dataset using RoBERTa large. We test on validation data split off from each respective training set, which allows us to measure how much hypothesis-only information is present within each dataset. The static adversarial and dynamic adversarial datasets have the lowest hypothesis-only accuracy. To measure high-overlap entailment instances, we find examples where the hypothesis has high~(>90\%) word overlap with the premise and compute how often the label is entailment. Such examples appear less frequently in DADC data.

% \paragraph{Generalization Versus Train-test Overlap} Given that we are collecting nearly 1000 examples per premise for each data collection method, there will inevitably be some overlap in the train and test sets. One concern with the more diverse DADC data is that it is not actually leading to better model generalization, but rather the data simply has more train-test overlap. To study this, we first analyze the amount of train-test overlap by taking each sentence in the test set and finding its most similar sentence in the train set. In the top of Figure~\ref{fig:overlap}, we plot a histogram of these different Inter-BLEU scores for the three different data collection methods.
\section{Related Work}

\noindent \textbf{Post-hoc Adversarial Filtering.} In adversarial filtering~\cite{lebras-etal-2020-adversarial,zellers-etal-2018-swag}, one takes an existing dataset and trains a model on the most difficult subportion of the data. Adversarial filtering shares motivations with adversarial data collection---difficult examples are more informative for learning---but it is focused on post-hoc data filtering rather than collection of new data.\smallskip %Moreover, in DADC we train on all of the collected examples, whereas adversarial filtering purposefully deletes easy examples.

\noindent \textbf{Adversarial Training.} Rather than having humans craft adversarial inputs, past work automatically generates adversarial examples and trains on them~\cite{goodfellow2014explaining,ribeiro2018semantically,ebrahimi2017hotflip}. The main downside of such approaches is their limited diversity---they focus on specific aspects like paraphrase~\cite{ribeiro2018semantically} or syntax~\cite{iyyerscpn2018} whereas DADC examples are only limited by human creativity.\smallskip

\noindent \textbf{Active Learning.} Active learning~\cite{lewis1994sequential}, especially when performed using an uncertainty-based acquisition function, is also closely related to DADC. The key differences are in the setup: active learning assumes access to unlabeled inputs, whereas in our setting we are interested in building datasets from scratch.\smallskip

\noindent \textbf{Other Data Quality Improvements.} Aside from adversarial data collection, researchers have explored numerous methods for improving data quality when using crowdsourcing. This includes feedback from experts~\cite{parrish2021does,nangia-etal-2021-ingredients}, gamifying the data collection process~\cite{yang-etal-2018-mastering,eisenschlos-etal-2021-fool}, encouraging counterfactual examples~\cite{kaushik-etal-2020-learning,gardner-etal-2020-evaluating}, or providing prompts that workers can edit~\cite{bowman-etal-2020-new,vania-etal-2020-asking}. Many of the ideas from these methods can be combined with adversarial data collection.
\section{Conclusion and Future Work}

We investigated dynamic adversarial data collection in the limit---over a large number of rounds until model performance starts plateauing---and demonstrated that data collected via this method is more valuable for training than alternatives, both on validation data and an expert-curated test set. We analyzed the collected data, showing that DADC yields examples that are more diverse, more complex, and contain fewer annotation artifacts compared to non-adversarial examples. Our results show that when building large training sets for training NLP models, data collected in an adversarial fashion with a continually updating model-in-the-loop can be more useful than standard model-agnostic collection in the long term.

In future work, it is vital to conduct similar experiments on different tasks, e.g., question answering and sentiment analysis, as well as on a larger number of contexts for NLI. Such experiments can provide insight into the generalizability of our findings. Moreover, given that a core benefit of DADC is promoting diversity and complexity of examples, one could explore other diversity-promoting methods of data collection. Lastly, our DADC setup is relatively simplistic in that we use a single target model and provide no other guides to the annotator; it would be interesting to provide generative models, model interpretations, or other methods to potentially further improve our DADC results.
\section*{Addressing Possible Ethical Concerns}

The premises that we use are sourced from publicly available sources and were vetted to ensure they contained no overtly offensive content. 
As described in main text, we designed our incentive structure to ensure that crowdworkers were well compensated (i.e., paid over minimum wage in the U.S.).
Our datasets focus on the English language as it is spoken in the United States. They are not collected for the purpose of designing NLP applications but to conduct a scientific study into collecting data for training machine learning models. 
We share our datasets to allow the community to replicate our findings and do not foresee any risks associated with the free use of this data.
\section*{Acknowledgements}

We thank Max Bartolo, Yixin Nie, Tristan Thrush, Pedro Rodriguez, and the other members of the Dynabench team for their valuable feedback on our crowdsourcing platform and paper. We also thank Alicia Parrish, Max Bartolo, Jessy Lin, and Nelson Liu for help annotating our test set.
\typeout{}
\bibliography{journal-abbrv,bib, anthology}
\bibliographystyle{acl_natbib}

\clearpage
\appendix
\section{Dataset Examples}\label{appendix:contexts}

Table~\ref{tab:contexts} shows the ten paragraphs that are used as the premises in our experiments.

\begin{table*}[!t]
\centering
\mycfs{7.5}
\begin{tabular}{p{15.5cm}}
\toprule
\textbf{Premises} \\
\midrule
{Sound is due to the vibrations of objects. A piano string produces sound because of its vibration when struck, or pulled to one side and then released. This vibration sets the air in rapid motion, and the result is the recording of the sound on our ear-drums. In old telephones, this recording corresponds to a film of sheepskin or bladder drawn over a hollow cup or cylinder. When the head of a drum is struck with a small stick it vibrates. In this case the vibrations are set in motion by the blow, while in the telephone a similar phenomenon is the result of vibratory waves falling from the voice on the thin membrane, or disk of metal, in the transmitter. When these vibrations reach the ear-drumm the nervous system, corresponding to electricity in the mechanical telephone, carries this sound to our brains where it is recorded and understood. In the telephone the wire, charged with electricity, carries the sound from one place to another.} \\
\midrule
{Michael Faraday was born at Newington, Surrey, on September 22, 1791, and was the third of four children. His father, James Faraday, was the son of Robert and Elizabeth Faraday, of Clapham Wood Hall, in the north-west of Yorkshire, and was brought up as a blacksmith. He was the third of ten children, and, in 1786, married Margaret Hastwell, a farmer's daughter. Soon after his marriage he came to London, where Michael was born. In 1796 James Faraday, with his family, moved from Newington, and took rooms over a coach-house in Jacob's Well Mews, Charles Street, Manchester Square. In looking at this humble abode one can scarcely help thinking that the Yorkshire blacksmith and his little family would have been far happier in a country house than in their new crowded London one, however, had he remained in the countryside, it is difficult to see how the genius of young Michael could have met with the requisites for its development. }\\
\midrule
{I had demonstrated by repeated experiments that inoculations of yellow fever blood into animals--dogs, rabbits, guinea pigs--gives a negative result. However, this negative result might be because these animals are not susceptible to the disease. In the civil hospital in Vera Cruz in 1887, Dr. Daniel Ruiz ran a single inoculation experiment on a man. But, this experiment was inconclusive because the patient from whom the blood was obtained was in the eighth day of the disease, and it was quite possible that the specific germ was destoyed at that point. These were the facts surrounding yellow fever when Dr. Reed and his associates commenced their investigations in Cuba during the summer of 1900. In a preliminary note read at the meeting of the American Public Health Association, October 22, 1900, the board gave a report of three cases of yellow fever which they believed to be direct results of mosquito inoculations. }\\
\midrule
{There are other signs of a coming change in the weather known less generally. When birds of long flight, such as swallows and others, hang about home and fly low—rain or wind may be expected. Also when animals seek sheltered places, instead of spreading over their usual range: when pigs carry straw to their sties; and when smoke from chimneys does not ascend readily, an unfavourable change may be looked for. Dew, on the other hand, is an indication of fine weather. So is fog. Neither of of these two formations occurs under an overcast sky, or when there is much wind.} \\
\midrule
{A fierce onslaught was made against Alvinczy's position by Massena's corps. It was entirely unsuccessful, and the French were repulsed with the serious loss of three thousand men. Bonaparte's position was now even more critical than it had been at Castiglione; he had to contend with two new Austrian armies, one on each flank, and Wurmser with a third stood ready to sally out of Mantua in his rear. If there should be even partial cooeperation between the Austrian leaders, he must retreat. But he felt sure there would be no cooeperation whatsoever. }\\
\midrule
{The pendulum had swung---it was no longer the Federalist merchants of New England who were discontent with the policies of the governement, but the planters of the South and particularly of South Carolina. New England was now in favor of a protective tariff. Webster, New England's foremost man at Washington, had voted against the tariff of 1816, but had changed his mind and supported a higher tariff in 1824, and a still higher in 1828. The planters of the South had not found it easy to manufacture goods. They had little or nothing, therefore, to protect against the products of European countries. On the contrary, they exported much to England, and imported from England and other countries many of the things they consumed. Accordingly, they were opposed to the whole system of tariff taxation and desired free trade.} \\
\midrule
{The water was wide and deep, so that he could not cross it. He, however, went down to the brink of the water, and got a good drink. This refreshed him very much, and then he went back again up the bank, and lay down upon the grass there to rest. Presently two cows came down to the water, on the side opposite to where Tony was sitting.} \\
\midrule
{The death of Socrates was brought under three of his enemies---Lycon, Meletus, and Anytus, the last a man of high rank and reputation in the state. Socrates was accused by them of despising the ancient gods of the state, introducing new divinities and corrupting the youth of Athens. He was charged with having taught his followers, young men of the first Athenian families, to despise the established government, to be turbulent and seditious, and his accusors pointed to Alcibiades and Critias, notorious for their lawlessness, as examples of the fruits of his teaching.} \\
\midrule
In some places the wires came very near together, and in others the spaces between them were so wide, that Wallace thought that the squirrel, if by any chance he should ever get put into the cage, would be very likely to squeeze his way out. Then, besides, Phonny had not measured his wires in respect to length, but had cut them off of various lengths, taking care however not to have any of them too short. The result was that the ends of the wires projected to various distances above the board, presenting a ragged and unworkmanlike appearance. \\
\midrule
Garrity was the most sinister figure in organized baseball. Once a newspaper reporter, he had somehow obtained control of the Rockets by chicanery and fraud. Sympathy and gratitude were sentiments unknown to him. He would work a winning pitcher to death, and then send the man shooting down to the minors the moment he showed the slightest symptom of weakness. He scoffed at regulations and bylaws; he defied restraint and control; he was in a constant wrangle with other owners and managers; and as a creator of discord and dissension he held the belt. \\
\bottomrule
\end{tabular}
\vspace{-0.2cm}
\caption{The ten paragraphs we use as premises in our experiments. We refer to these contexts as Sound, Faraday, Yellow, Weather, Battle, Tariff,
Water, Socrates, Wires, and Garrity, respectively.}
\label{tab:contexts}
\end{table*}

\section{Mechanical Turk Interface}\label{appendix:turk}

\begin{figure*}[t]
\centering
\includegraphics[trim={0.0cm 0cm 0.0cm 0.0cm},clip, width=1.0\textwidth]{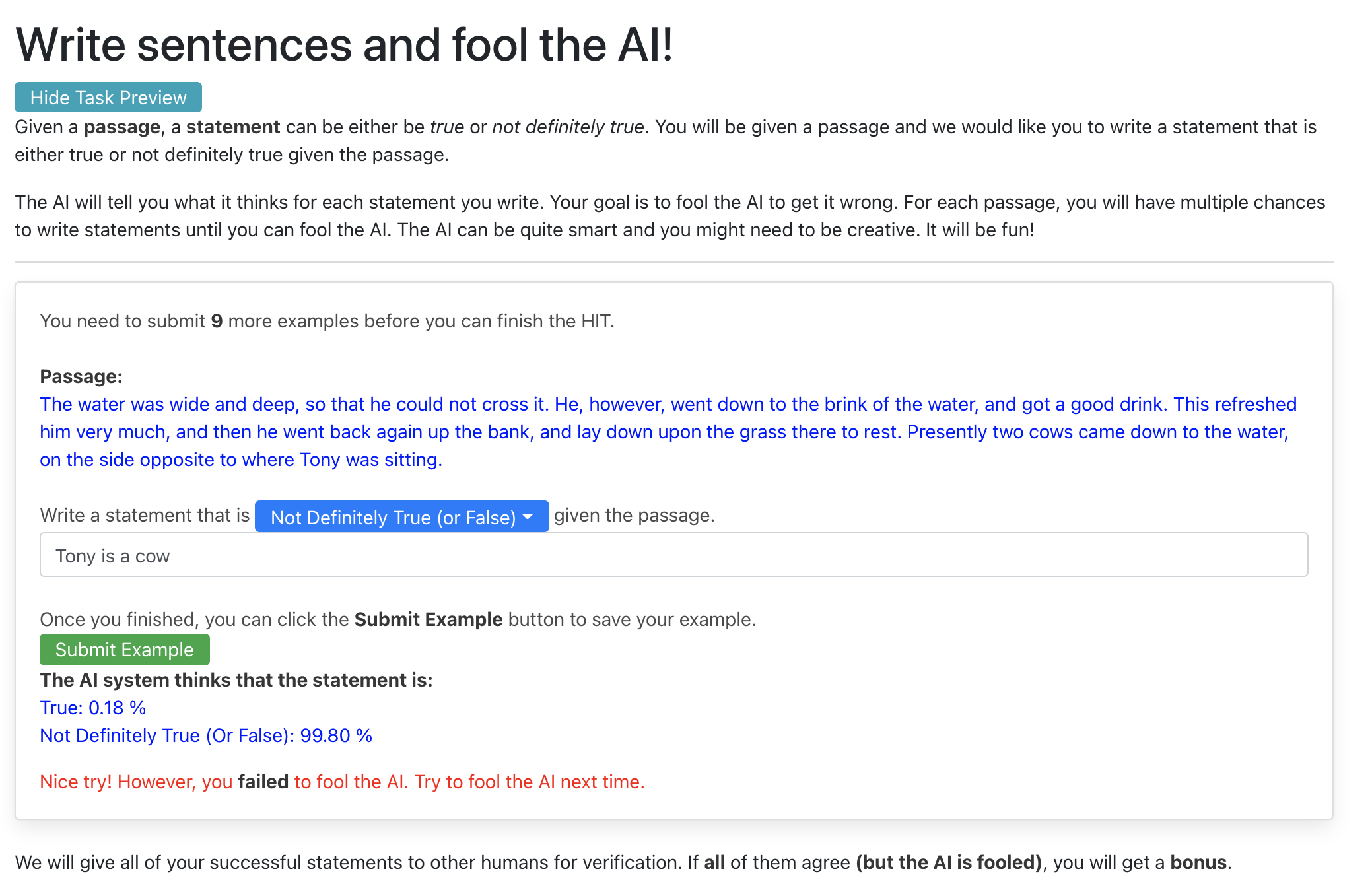}
\vspace{-0.6cm}
\caption{Our Amazon Mechanical Turk interface for the model-in-the-loop setting.}
\label{fig:interface}
\end{figure*}

Figure~\ref{fig:interface} shows our Amazon Mechanical Turk interface for the model-in-the-loop setting.

\end{document}